\documentclass[11pt]{article}

\usepackage{acl}
\usepackage{times}
\usepackage{latexsym}
\usepackage[T1]{fontenc}
\usepackage[utf8]{inputenc}
\usepackage{microtype}
\usepackage{inconsolata}
\usepackage{graphicx}

\usepackage{tabularx}
\usepackage{booktabs}
\usepackage{multirow}
\usepackage{subcaption}

\title{Wiktionary as a Crowdsourced Lexicon for English Dialects}

\author{Sidney Wong \\
  Centre for Sustainability Research, University of Otago \\
  Te Pūnaha Matatini Centre of Research Excellence for Complex Systems, New Zealand \\
  \texttt{sidney.wong@otago.ac.nz}}

\begin{document}
\maketitle
\begin{abstract}
    This paper evaluates Wiktionary as an ethically crowdsourced lexicon for English dialects. We took a two-phase approach, providing an in-depth descriptive analysis of the crowdsourced lexicon for 12 national varieties of English before applying the lexicon to geo-referenced, country-level social media language data to examine the real-world performance of this crowdsourced dialect lexicon. We demonstrate that Wiktionary matches or exceeds the coverage of traditional dictionaries, such as the Oxford English Dictionary (OED), for regional and Outer-Circle varieties. Our dialect-specific case study on New Zealand English found high alignment between Wiktionary and the OED based on word-formation patterns (R = 0.883). Similarly, we observed high alignment between the dialect lexicon and geo-referenced social media language. While this paper found that Wiktionary has broad coverage of lexical properties, it also highlighted some of the macro-challenges involved in evaluating dialect-responsive language resources and tools, such as the role of language contact in dialects and register effects in web-based corpora.
\end{abstract}

\section{Background and Motivation}
\label{sec:1}

    Despite recent advances in transformer-based large language models \cite{faisal_testing_2025}, an urgent need remains to develop dialect-responsive natural language processing (NLP) tools and resources. This is particularly challenging in low-resource language contexts, such as closely related varieties or dialects \cite{zampieri_natural_2020}. A key reason for this limitation is the over-reliance on web corpora (including social media) during system development \cite{hovy_five_2021}. Consequently, for pluricentric languages such as English, French, Portuguese, and Spanish, contemporary systems remain prone to geographic bias \cite{faisal_geographic_2023}.
    
    With the popularisation of Web-as-Corpus approaches \citep{kilgarriff_introduction_2003}, one approach to improving dialect responsiveness has been a shift towards more geographically representative corpora derived from social media language data \cite{dunn_geographically-balanced_2020}. This is because the metadata associated with these sources can be treated as ground truth for geographic dialect variation \cite{johnson_geography_2016}. However, geographic information does not necessarily align with real-world linguistic behaviour \cite{sloan_who_2015}. Other forms of geographic bias, such as non-local bias \cite{dunn_measuring_2020}, may further reduce representation within corpora.
    
    Representative corpora are only one approach to improving fairness and equitable access in NLP systems. There is also a demand for resources and tools to evaluate dialect responsiveness. Dialect lexicons (including dictionaries and textual corpora) play a vital role in developing dialect-responsive language technologies \cite{joshi_natural_2025}. For example, lexicons can be used to address the out-of-vocabulary (OOV) problem, especially in low-resource varieties or dialects \cite{joshi_natural_2025}. With this in mind, we argue that ethically acquired open-source language resources, such as Wiktionary\footnote{\href{https://www.wiktionary.org/}{https://www.wiktionary.org/}}, provide essential ground-truth lexical information about dialects in pluricentric language contexts such as English.

\subsection{Lexicons and Dictionaries}

    At the most foundational level, the lexicon comprises the basic building blocks of language. A lexicon is not restricted to words alone, but also includes affixes, stems, and more complex constructions such as phrasal units or idioms \citep{jackendoff_whats_2002}. The creation and curation of lexicons remain core activities in dialectology \citep{chambers_dialectology_1998} and language documentation and conservation \citep{haviland_documenting_2006}. Some of the most well-known lexicons are dictionaries, which are reference works used for text reception, text production, or translation and were developed in response to the mass production of books and other written media \citep{bergenholtz_what_2012}. The most prominent example of a dictionary is the \textit{Oxford English Dictionary} (OED), which, in its online form, amasses the meaning, history, and usage of over 500,000 words and phrases in English\footnote{\href{https://www.oed.com/}{https://www.oed.com/}}. Institutional lexicons are by no means complete, with the primary challenges being size restrictions and the underspecification of lexical items \citep{klein_lexicology_2015}.
    
    Lexicons relating to non-standard' dialects (or varieties) may risk exclusion, as they are often omitted from dictionaries concerned with standard' language usage. Data collection efforts often rely on questionnaires and interviews with informants \citep{chambers_dialectology_1998}. Naturally, the resulting lexicons can vary in size and quality. Using New Zealand English (NZE) as an example, the now-defunct New Zealand Dictionary Centre held a database of approximately 42,000 entries \citep{bardsley_lexicography_2009}. However, this figure does not reflect actual usage. More contemporary sources, such as OED Online, include 1,404 entries coded to NZE alone. This is in sharp contrast to earlier sources, which combined New Zealand and Australian index entries into a lexicon of 820 entries \citep{turner_english_1966}. The exclusion of some lexical properties, such as \textit{listemes} \citep{di_sciullo_definition_1987}, also poses a challenge in lexicography because of the focus on individual lexical units \citep{klein_lexicology_2015}.
    
\subsection{Web-as-Corpus and Online Dictionaries}

    In the same way that the printing press played a transformational role in disseminating lexicons \cite{tarp_concept_2017}, the internet has provided a new medium for the development of lexicons \cite{carr_internet_1997}. As social media has become a major source of geo-referenced language data, computational approaches have increasingly been used to extract lexical dialect features \citep{nguyen_dialect_2021}. \citet{grieve_analyzing_2017} documented the emergence of innovative lexical features on Twitter (now X) by measuring changes in relative frequency over time. Using a similar methodology, \citet{mahler_lexical_2020} applied this approach to Reddit data. More recently, statistical language modelling has been used to extract lexical dialect features \citep{xie_extracting_2024}.
    
    Online dictionaries, such as Urban Dictionary and Wiktionary, have also provided another avenue for examining lexical information \citep{johns_mining_2019}. Wiktionary is a multilingual, web-based dictionary collaboratively edited by users \citep{meyer_wiktionary_2012}. Similar to Wikipedia, Wiktionary is a form of digital commons. Unlike published dictionaries, which are curated by lexicographers, Wiktionary relies on registered volunteers to create individual entries. Users contribute the etymology, pronunciation, definitions, usage notes, derived forms, related terms, alternative forms, and translations for each entry. Wiktionary's crowdsourced nature also comes with limitations. Because volunteers can edit entries, their quality may vary, and they remain susceptible to errors and misinformation. In one notable example, nearly half of the articles in the Scots-language Wikipedia were authored by a non-Scots speaker, which raises serious concerns about the quality of crowdsourcing \citep{brooks_shock_2020}.

\subsection{Lexicons for NLP}

    Annotated corpora, lexicons, and grammars played a crucial role in the early development of NLP systems \cite{laporte_lexicons_2009}. Lexicons provide not only morphosyntactic information but also syntactic-semantic information. One of the first lexicons for English, WordNet \citep{miller_wordnet_1995}, provided semantic relations between words. Other lexicons providing syntactic-semantic properties of verbs include FrameNet \citep{fillmore_starting_1994}, VerbNet \citep{kipper_class-based_2000}, and COMLEX \citep{grishman_comlex_1994}. As a form of machine-readable structured data \citep{ylonen_wiktextract_2022}, Wiktionary is a valuable source of lexical-semantic information \citep{zesch_extracting_2008}. With the advent of vector-based word embedding models \citep{mikolov_efficient_2013}, Wiktionary-derived embeddings were used to adapt models to multilingual contexts \citep{de_melo_wiktionary-based_2015}.
    
    The demand for lexicons such as WordNet has reduced since the shift towards transformer-based LLMs \citep{meconi_large_2025}. Similarly, LLMs are now capable of producing fit-for-purpose dictionaries and lexicons for text reception, machine translation, and text production \citep{de_schryver_generative_2023}. However, this does not mean that the need for lexicons in the development of NLP systems no longer exists \citep{lew_dictionaries_2024}. As mentioned earlier, the OOV problem remains a significant issue in low-resource language contexts \citep{joshi_natural_2025}. Even in state-of-the-art LLMs, a high concentration of OOV items can lead to a significant reduction in performance \citep{balde_evaluation_2025}. Similarly, standard tokenisation and instruction tuning do not improve the performance of LLMs on reasoning tasks for 'non-standard' varieties of English \citep{lin_assessing_2025}.

\subsection{Summary}

    The upkeep of labour-intensive dialect dictionaries and reference corpora is often unsustainable for low- or under-resourced language varieties. In the case of the New Zealand Dictionary Centre, operations ceased in 2012 \citep{bardsley_lexicography_2009}. This is a major challenge as lexis is the most salient distinguishing feature of NZE \citep{deverson_handling_2000}. While dialect lexicons may take years to curate \citep{klein_lexicology_2015}, an advantage of Wiktionary as a source of lexical dialect information is its near-contemporaneous nature \citep{meyer_wiktionary_2012}. This is particularly important in the context of emergent internet-origin lexical dialect features \citep{mahler_lexical_2020}. Though there has been limited research on English dialects, Wiktionary has been applied to the dialects of other language contexts, including Germanic \citep{bernhard_adding_2014}, Sinitic \citep{chang_wikihan_2022}, and Romance languages \citep{sprugnoli_ciallabacialla_2025}.
    
    The availability of an Application Programming Interface (API) enables the ethical acquisition of Wiktionary entries. This is particularly important for language tools and resources, such as digital corpora, to be consistent with FAIR (Findability, Accessibility, Interoperability, and Reusability) principles \citep{frey_fair_2020}. With geographic dialect bias remaining a significant issue in the development of language models \citep{faisal_testing_2025}, there is a critical need to develop structured lexicons and dictionary-based features for underserved dialects to build fair and equitable NLP applications \citep{joshi_natural_2025}. To evaluate the viability of Wiktionary as a crowdsourced dialect lexicon, this study addresses two central research questions:
    
    \begin{itemize}
        \item[RQ1] How does Wiktionary's regional English coverage compare to traditional lexicons such as the OED?
        \item[RQ2] Can Wiktionary-derived features identify regional dialect usage across social media text registers?
    \end{itemize}

\section{Methodology}

    To address our research questions, we begin by comparing crowdsourced dialect lexicons from Wiktionary with existing published lexicons, namely the OED. We then evaluate these crowdsourced dialect lexicons against geo-referenced language data. Throughout our analysis, we contextualise the results and findings with reference to NZE. As one of the newest, most geographically isolated, and most well-studied Inner-Circle varieties of English \citep{hay_origins_2008}, NZE is particularly well suited to exploring the efficacy of NLP tools and resources in a low-resource context. Although 96\% of the NZE lexicon is shared with other Inner-Circle varieties \citep{deverson_handling_2000}, lexis remains the most distinctive feature of NZE, characterised by a high degree of borrowings from te reo Māori as a result of language contact through settler colonialism.

\subsection{Data Sources}

    One of the primary goals of this paper is to examine Wiktionary as an open-source dialect lexicon and as an alternative to institutional dictionaries, such as the Oxford English Dictionary (OED). In this section, we provide an overview of Wiktionary, as well as Reddit, which serves as our source of geo-referenced social media language data.

\subsubsection{Wiktionary: The Free Dictionary}

    \begin{table}
        \centering
        \footnotesize
        \begin{tabularx}{\columnwidth}{l *{1}{>{\centering\arraybackslash}X}l}
            \toprule
            Category & $n$ & Example \\
            \midrule
            African & 13 & Botswanan, Cameroonian \\
            Antarctic & - & - \\
            Asian & 3 & East Asian, South Asian \\
            Caribbean & 8 & Bahaman, Barbadian \\
            Commonwealth & 16 & Australian, Bangaldeshi \\
            European & 6 & British, Channel Islands \\
            Middle Eastern & 3 & Egyptian, Israeli \\
            North American & 7 & American, Bermudian \\
            Oceanian & 8 & Fijian, French Polynesian \\
            \bottomrule
        \end{tabularx}
        \caption{\label{tab:wiktionary_categories} Summary table of the Wiktionary categories of regional varieties including the number of subcategories ($n$) and examples.}
    \end{table}

\begin{table*}[t]
    \centering
    \footnotesize
        \begin{tabularx}{\linewidth}{lll *{5}{>{\centering\arraybackslash}X}}
            \toprule
            \multirow{2}{*}{Grouping} & \multirow{2}{*}{Country} & \multirow{2}{*}{Subreddit} & \multirow{2}{*}{Members} & \multicolumn{2}{c}{Submissions} & \multicolumn{2}{c}{Comments} \\
            & & & & $n$ & $\bar{x}$ & $n$ & $\bar{x}$ \\
            \midrule
            Outer-Circle & India & \texttt{r/india} & 3.2M & 481K & 64.5 & 12.5M & 202.8 \\
            Outer-Circle & Pakistan & \texttt{r/pakistan} & 575K & 87K & 56.9 & 2.4M & 212.0 \\
            Outer-Circle & Malaysia & \texttt{r/malaysia} & 1.3M & 73K & 55.3 & 3.5M & 179.0 \\
            Outer-Circle & Philippines & \texttt{r/Philippines} & 3.3M & 331K & 52.0 & 13.6M & 138.7 \\
            Outer-Circle & Kenya & \texttt{r/Kenya} & 219K & 24K & 42.3 & 475K & 150.6 \\
            Outer-Circle & South Africa & \texttt{r/southafrica} & 362K & 56K & 51.5 & 1.6M & 212.2 \\
            Inner-Circle & Canada & \texttt{r/canada} & 4.2M & 175K & 67.5 & 18.5M & 241.3 \\
            Inner-Circle & United States & \texttt{r/usa} & 119K & 23K & 54.8 & 102K & 293.0 \\
            Inner-Circle & Australia & \texttt{r/australia} & 2.7M & 198K & 60.0 & 12.9M & 215.1 \\
            Inner-Circle & New Zealand & \texttt{r/newzealand} & 794K & 136K & 56.2 & 8.7M & 198.7 \\
            Inner-Circle & Ireland & \texttt{r/ireland} & 1.2M & 235K & 53.3 & 10.5M & 178.3 \\
            Inner-Circle & United Kingdom & \texttt{r/unitedkingdom} & 5.4M & 146K & 67.3 & 7.4M & 236.7 \\

            \bottomrule
        \end{tabularx}
    \caption{Summary table of Reddit communities including the associated national variety of English (Country), the name of the community (Community), the number of subscribers  as of June 2025 (Members), and the total number of words ($n$) and mean word length ($\bar{x}$) for Submission Posts and Comments.}
    \label{tab:corpus_summary}
    \end{table*}

    As of December 2024, the English edition of Wiktionary contained over 8.245 million entries. Of interest to the present study are the bespoke categories comprising entries associated with different language varieties. Entries are labelled with metadata that is then used to generate distinct categories. For English, entries are labelled as regional varieties and organised into broader geographic categories (\textit{Oceanian English}) and subcategories (\textit{New Zealand English}). We provide a summary of these categories in Table \ref{tab:wiktionary_categories}. Some categories, such as \textit{American English}, are further subdivided into subnational (\textit{Midwestern English}, \textit{Appalachian English}) or social (\textit{African-American English}, \textit{American slang}) varieties or dialects. Of interest to our analysis are the subcategories that correspond to national varieties of English. 
    
    National varieties can be classified as either Inner-Circle or Outer-Circle varieties \citep{kachru_standards_1985}. The distinction between the Inner-Circle and Outer-Circle varieties reflects two distinct waves of British colonial expansion which shapes how these varieties are perceived, institutionalised, and used today \citep{kachru_standards_1985}. The Inner Circle refers to varieties associated with countries where English is the majority language. Meanwhile, Outer-Circle varieties refer to those associated with former British colonies. We selected a geographically diverse sample of six Inner-Circle (Western Europe, North America, and Oceania) and six Outer-Circle (South Asia, Sub-Saharan Africa, and South East Asia) national varieties or dialects. One limiting factor was that this sample had to correspond with existing country-level communities on Reddit which we discuss in the next section.

\subsubsection{Other Data Sources}

    In addition to Wiktionary entries, we use Reddit language data to validate our crowdsourced dialect lexicon. As a source of real-world dialect usage, we use implicit geographic information (such as place-based digital communities) as a proxy for geographic dialect variation (\citealp{hamre_geographic_2024}; \citealp{wong_language_2026}). Our primary data source comes from the Pushshift Dumps, an ongoing data collection effort that captures the top 40,000 Reddit communities via the Pushshift API \citep{baumgartner_pushshift_2020}. We restrict our sample to country-level communities associated with twelve national varieties of English. These correspond with six Inner-Circle and six Outer-Circle varieties based on the Three Circles of English model \citep{kachru_standards_1985}. Corpus characteristics are provided in Table \ref{tab:corpus_summary}.

    One consideration is how Reddit functions as a register \citep{biber_register_2009}. Reddit does not consist of a single text type but instead serves multiple communicative purposes. With this in mind, we compare two contrasting text types: Submission Posts and Comments. Users generally use Submission Posts to pose questions or seek advice from the community, whereas Comments mostly originate from community subscribers responding to posts. To illustrate the differences between these text types, the mean word count of Submission Posts is much lower ($\bar{x}$; Outer-Circle $=58.1$, Inner-Circle $=60.2$) than that of Comments ($\bar{x}$; Outer-Circle $=175.0$, Inner-Circle $=217.1$).
    
\subsection{Analytical Pipeline}

    We take a two-phase approach to address the research questions. The results from Phase 1, Descriptive Analysis, are used to address RQ1, while the results from Phase 2, Corpus Validation, are used to address RQ2.

\paragraph{Phase 1: Descriptive Analysis}

    To answer RQ1, we take a descriptive approach by comparing the coverage of lexical dialect features in the OED and Wiktionary. First, we compare the coverage rates of Wiktionary and the OED for each national dialect variety. Second, we manually code the Wiktionary entries for NZE based on word-formation processes and cross-validate them against the OED. The Wiktionary entries for NZE offer a suitable dialect-specific case study because of their manageable size ($n=1,415$). The coded information includes word-formation processes, Wiktionary definitions, stem or root words, morphological processes, usage notes, and other varieties that share the lexical feature.

\paragraph{Phase 2: Corpus Validation}
    
    At this stage, we introduce the geo-referenced language data from Reddit. We use the same evaluation method as \citet{grieve_analyzing_2017}, comparing relative word frequencies. This is calculated as the rate of occurrence per 100,000 words ($\mu_{i}$). This measure allows us to adjust for overall sample size and account for the varying sizes of each country-level community. More specifically, this phase enables us to evaluate and validate the effectiveness of the crowdsourced dialect lexicon using real-world language data.

\subsection{Evaluation}

    The results from our analysis will be largely descriptive. As a relative measure of word frequency, $\mu_{i}$ allows us to directly compare the occurrence of lexical dialect features across corpora. As an example, \textit{skux} (`stud, player, womaniser') should have the greatest $\mu_{i}$ in \texttt{r/newzealand} as feature that originated from New Zealand English. Although we do not conduct statistical modelling to test hypotheses, in Phase 2 we predict strong alignment between Wiktionary entries and subreddit communities and expect the highest mean relative frequencies of lexical dialect features to occur in their corresponding national subreddits.

    \begin{table}[t]
        \centering
        \footnotesize
        \begin{tabularx}{\columnwidth}{l*{3}{>{\centering\arraybackslash}X}}
            \toprule 
            Dialect & OED & Wiki & $\%\Delta$ \\ 
            \midrule
            Malaysian & 19 & 450 & 2,268.4 \\
            Filipino & 93 & 1,493 & 1,505.4 \\
            Canadian & 550 & 2,834 & 415.3 \\
            Indian & 721 & 1,816 & 151.9 \\
            Irish & 1,145 & 2,105 & 83.8 \\
            Australian & 2,435 & 3,381 & 38.9 \\
            New Zealand & 1,435 & 1,415 & -1.4 \\
            South African & 982 & 928 & -5.5 \\
            American & 15,923 & 11,899 & -25.3 \\
            British & 16,191 & 11,283 & -30.3 \\
            \midrule
            Kenyan & - & 65 & - \\
            Pakistani & - & 132 & - \\
            \bottomrule
        \end{tabularx}
        \caption{\label{tab:wiktionary} Summary table comparing the total number of entries from the OED and Wiktionary (Wiki) by national variety (Dialect).}
    \end{table}

    \begin{figure}[t]
        \centering
        \includegraphics[width=\linewidth]{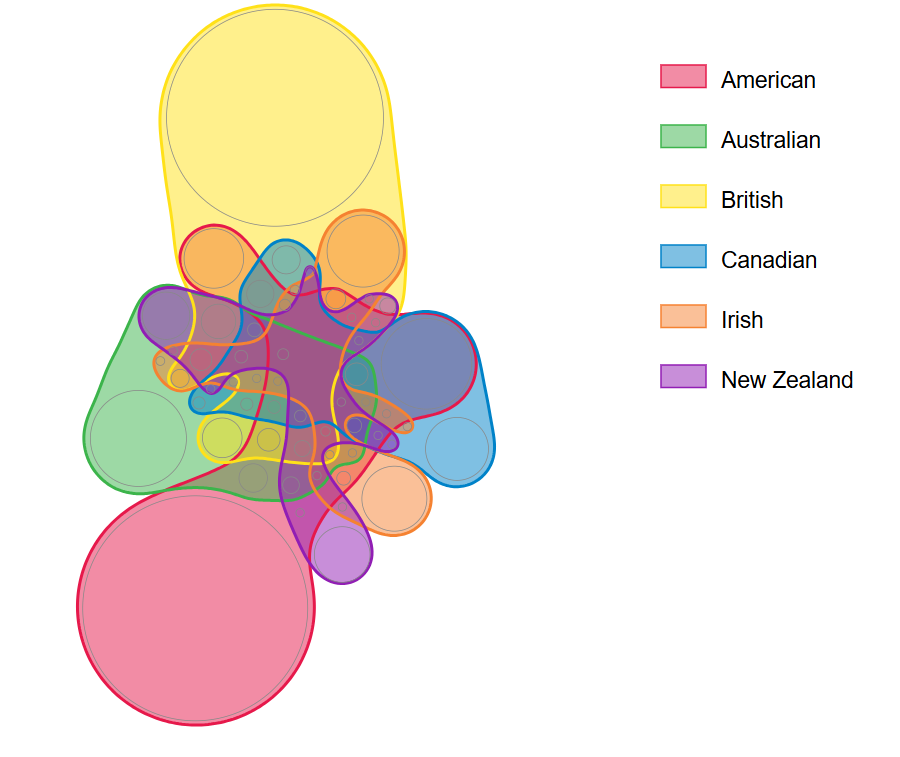}
        \caption{\label{fig:euler_wiktionary} Euler diagram illustrating the degree of overlap in Wiktionary entries for the six Inner-Circle varieties of English. Similar to Venn diagrams, a Euler diagram visualises sets and their relationships particularly those with complex hierarchies.}
    \end{figure}

\subsection{Data Processing}

    An important consideration for crowdsourced corpora is data quality \citep{schafer_good_2013}. We retrieved the entries as a \texttt{.json} file through the Wiktionary API using Python's \texttt{urllib.request} module. We then filtered out entries longer than three words, which approximately corresponds to trigrams. As an additional processing step, we removed duplicates by aggregating entries. We applied a similar processing procedure to the geo-referenced language data, first removing punctuation and special characters (excluding `\&' and `\$') from each observation string before converting all uppercase characters to lowercase. Of particular relevance to NZE is the variable use of diacritics to represent long vowels in te reo Māori orthography \citep{deverson_new_1991}: with a \textit{tohutō} (`macron'), as two consecutive vowels, or with no diacritic. For the purposes of this paper, we did not standardise the Wiktionary entries. Once we had cleaned the observation strings, we joined them into a single string, which was then split into unigrams, bigrams, and trigrams using \texttt{nltk.ngrams()} from the \texttt{nltk} package. We then used the \texttt{Counter} class from the \texttt{collections} module to produce a dictionary object of key-value pairs, which we converted into a Pandas \texttt{DataFrame} for easier manipulation.

\subsection{System Requirements}

    The analysis was carried out in Google Colab using Python with the Pandas, NumPy, and NLTK packages, and bar graphs were generated using Matplotlib. Euler diagrams were generated in R using the \texttt{eulerr} package. 

\subsection{Data Availability}

    The processed Reddit language data covering six Inner-Circle and six Outer-Circle varieties are available in the Hugging Face repository Reddit Corpus of Global Language Use (ReCGLU)\footnote{\url{https://doi.org/10.57967/hf/8197}}. All associated code and output datasets (in .csv format) are hosted in the paper's Hugging Face repository\footnote{\url{https://doi.org/10.57967/hf/10458}}.

\section{Results}

    A comparison of coverage as of December 2024 (Table \ref{tab:wiktionary}) reveals that, while the OED remains slightly larger overall (by 13.1\%; $n=37,679$ vs. $n=33,043$), the two resources capture distinct dimensions of Inner-Circle English. The OED retains higher coverage of American and British English (exceeding Wiktionary by 33.8\% and 43.5\%, respectively). In contrast, Wiktionary demonstrates superior coverage of regional Inner-Circle varieties, offering larger lexicons for Canadian, Irish, and Australian English. Regarding Outer-Circle Englishes, Wiktionary provides coverage for all six national varieties. By comparison, the OED contains entries for South African English, Indian English, Malaysian English, and Filipino English but lacks coverage of Kenyan English and Pakistani English. As shown in the Euler diagram in Figure \ref{fig:euler_wiktionary}, lexical sharing across Inner-Circle varieties aligns closely with regional and historical dialect clusters. The strongest cross-varietal overlap occurs between North American Englishes (American and Canadian, $n=1,476$) and British and Irish Englishes ($n=914$). In contrast, NZE exhibits a distinct Australasian/Oceanian alignment, sharing 400 entries exclusively with Australian English and a further 201 entries with both British and Australian English.

\subsection{Dialect-Specific Case Study}
\label{subsec:r1}

    \begin{figure}[t]
        \centering
        \includegraphics[width=\linewidth]{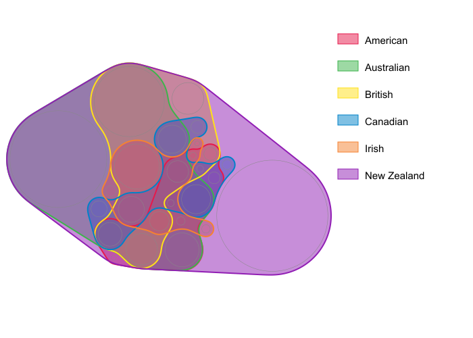}
        \caption{\label{fig:euler} Euler diagram illustrating the degree of overlap in Wiktionary entries filtered by NZE and the six Inner-Circle varieties of English.}
    \end{figure}

    \begin{table}
        \centering
        \footnotesize
        \begin{tabularx}{\columnwidth}{l *{3}{>{\centering\arraybackslash}X}}
            \toprule
            Word Formation Process & OED & Wiki & $\%\Delta$ \\
            \midrule
            Initialism & 3 & 14 & 366.7 \\ 
            Blend & 3 & 9 & 200.0 \\ 
            Backformation & 3 & 6 & 100.0 \\ 
            Unknown & 21 & 38 & 81.0 \\ 
            Proper name & 37 & 66 & 78.4 \\ 
            Borrowing & 287 & 399 & 39.0 \\ 
            Conversion & 80 & 80 & 0.0 \\ 
            Derivative & 230 & 228 & -0.9 \\ 
            Imitative & 8 & 6 & -25.0 \\ 
            Variant & 65 & 41 & -36.9 \\ 
            Shortening & 59 & 34 & -42.4 \\ 
            Compound & 587 & 335 & -42.9 \\
            \midrule
            Uncertain & 10 & - & - \\ 
            Arbitrary & 4 & - & - \\ 
            Acronym & - & 3 & - \\ 
            Idiomatic & - & 73 & - \\ 
            \bottomrule
        \end{tabularx}
        \caption{\label{tab:dictionary_compare} Summary table comparing the number of entries classified by word formation processes between the OED and Wiktionary (Wiki). The categories are ordered by percentage difference ($\%\Delta$) between OED and Wiktionary.}
    \end{table}

    \begin{table*}[t]
        \centering
        \footnotesize
        \begin{tabularx}{\linewidth}{l*{6}{>{\centering\arraybackslash}X}}
            \toprule 
            \multirow{ 2}{*}{Dialect} & 
            \multicolumn{3}{c}{Submission Posts} & 
            \multicolumn{3}{c}{Comments} \\ 
             & Community & \textbf{$\mu_{i}$} & Alignment & Community & \textbf{$\mu_{i}$} & Alignment \\
            \midrule
            British & \texttt{r/unitedkingdom} & 5,475.2 & Yes & \texttt{r/malaysia} & 3,973.3 & No \\
            Irish & \texttt{r/ireland} & 889.1 & Yes &\texttt{r/ireland} & 771.1 & Yes \\
            Australian & \texttt{r/australia} & 743.4 & Yes &\texttt{r/australia} & 333.4 & Yes \\
            New Zealand & \texttt{r/newzealand} & 829.5 & Yes & \texttt{r/Philippines} & 498.8 & No \\
            American & \texttt{r/usa} & 5,538.7 & Yes & \texttt{r/canada} & 3,582.3 & No \\
            Canadian & \texttt{r/canada} & 443.2 & Yes & \texttt{r/canada} & 211.9 & Yes \\
            \bottomrule
        \end{tabularx}
        \caption{\label{tab:relative_frequency} Summary table presenting the Subreddit communities (Community) with the greatest mean relative frequency ($\mu_{i}$) and alignment by Inner-Circle variety of English (Dialect) grouped by text-type: Submission Posts or Comments.}
    \end{table*}

    We now investigate Wiktionary's coverage of one Inner-Circle variety: NZE. We found that borrowings from te reo Māori in Wiktionary were included in a separate category: \textit{English terms borrowed from Māori}. This category contained 324 entries. The combined total of the two NZE-related categories was 1,541 entries, or 1,332 entries after applying the data-processing steps. When New Zealand English entries only, we observed a significant overlap with British and Australian entries in as shown in the Euler diagram (Figure \ref{fig:euler}). This was expected due to the historic relationship between the three varieties. Of the processed entries, only 498 were unique to NZE, with the majority being borrowings from te reo Māori ($n=312$) and the remainder being formed through other word-formation processes ($n=186$). We present a comparison between Wiktionary and the OED for lexical features classified according to their word-formation processes in the OED in Table \ref{tab:dictionary_compare}. There is a strong positive correlation between the numbers of entries for each type of word-formation process (Pearson's $R=0.883$; $R^2=0.780$) when we exclude the Uncertain, Arbitrary, Acronym, and Idiomatic categories. Furthermore, the number of entries from the additional category was comparable to the number of te reo Māori entries ($n=337$) in the OED.

\subsection{Corpus Validation}
\label{subsec:r2}

    We present the mean relative frequency ($\mu_{i}$) for each Inner-Circle variety of English and its prevalence in each country-level subreddit community in Table \ref{tab:relative_frequency}. For Submission Posts, we observed strong alignment between each country-level variety and its associated subreddit community. This result was expected. Even with a small subscriber base, we observed this relationship in \texttt{r/usa}. However, the relationship observed in Submission Posts did not hold for Comments. The only varieties of English that maintained a consistent relationship across both Submission Posts and Comments were Irish, Canadian, and Australian English. The prevalence of American English features in \texttt{r/canada} was expected, given that both are North American varieties of English. We visualised the mean relative frequencies ($\mu_{i}$) for Submission Posts and Comments in the Appendix (see Figures \ref{fig:title_lexical} and \ref{fig:comment_lexical}).

\section{Discussion}

    \begin{table}
        \centering
        \footnotesize
        \begin{tabularx}{\columnwidth}{l *{2}{>{\centering\arraybackslash}X}}
            \toprule
            Entry & Submissions & Comments \\
            \midrule
            \textit{box of birds} & - & 11 \\ 
            \textit{box of fluffies} & - & 14 \\ 
            \textit{in the dogbox} & - & 7 \\ 
            \textit{jack a dandy} & - & - \\ 
            \textit{pack a sad} & - & 39 \\ 
            \textit{suck the kumara} & 1 & 5 \\ 
            \textit{turn to custard} & - & 36 \\ 
            \textit{up the boohai} & - & 5 \\ 
            \textit{warrant of fitness} & 5 & 278 \\ 
            \bottomrule
        \end{tabularx}
        \caption{\label{tab:mwe_entries} Table listing the total number of multi-word expressions (excluding borrowings from te reo Māori) in \texttt{r/newzealand} from Submission Posts and Comments.}
    \end{table}

    The findings suggest that Wiktionary complements traditional curated sources such as the OED and offers advantages in accessibility and coverage, particularly for Outer-Circle varieties and regional loanwords that are often neglected by Western-centric institutions. Based on this finding, we observe that the OED maintains broader historical coverage of primary Inner-Circle standard varieties (US and UK English), while Wiktionary matches or significantly exceeds the OED in capturing regional Inner-Circle varieties (Canadian, Irish, and Australian English) and Outer-Circle varieties (Filipino, Indian, and Malaysian English). Furthermore, Wiktionary's entry structure closely reflects lexicographical word-formation patterns ($R = 0.883$), while offering broader coverage of multi-word listemes and contemporaneous loanwords, as demonstrated by our dialect-specific case study.

\paragraph{Broad Coverage of Lexical Properties}

    Coverage remains a limitation of dialect lexicons \citep{bergenholtz_what_2012}. When compared with historical dialect lexicons, only 121 entries appeared in \citet{turner_english_1966}, suggesting either that 85.24\% ($n=699$) were no longer in general usage or that the majority of those entries were more relevant to Australian English (of those 121, only 50 were distinctive to NZE). While the coverage of Wiktionary and the OED was comparable for most national dialects, the most prominent difference between them was Wiktionary's inclusion of idiomatic entries (e.g., \textit{box of birds}, \textit{good as gold}). Also known as complex compounds or listemes, these entries address a known limitation of institutional lexicons, as listemes are typically excluded or not treated as distinct headwords in dictionaries \citep{klein_lexicology_2015}. In NZE, some of these listemes function alongside the stative verb \textit{to be} (e.g., \textit{to be a box of birds}, \textit{to be good as gold}). We present the token frequencies of these multi-word expressions in Table \ref{tab:mwe_entries}. Of the nine entries, only two were not present in the OED (\textit{suck the kumara}, \textit{turn to custard}). Even though Wiktionary had more comprehensive coverage of idiomatic phrases in NZE, the OED better reflects the actual usage of these phrases.

\paragraph{Language Contact in Dialects}

    \begin{table}
        \centering
        \footnotesize
        \begin{tabularx}{\columnwidth}{l *{4}{>{\centering\arraybackslash}X}}
            \toprule
            \multirow{2}{*}{Entry} & \multicolumn{2}{c}{\texttt{r/newzealand}} & \multicolumn{2}{c}{\texttt{r/Philippines}} \\
            & $n$ & $\mu_{i}$ & $n$ & $\mu_{i}$ \\
            \midrule
            \textit{mo} & 1,902 & 0.6 & 1,319,868 & 403.9 \\ 
            \textit{pa} & 3,338 & 1.1 & 1,174,769 & 359.5 \\ 
            \textit{para} & 372 & 0.1 & 499,040 & 152.7 \\ 
            \textit{eh} & 17,914 & 5.9 & 435,229 & 133.2 \\ 
            \textit{po} & 857 & 0.3 & 309,069 & 94.6 \\
            \bottomrule
        \end{tabularx}
        \caption{\label{tab:wiktionary_compare} Summary table including the token frequency ($n$) and mean relative frequency ($\mu_{i}$) for the top five NZE Wiktionary entries grouped by Subreddit communities.}
    \end{table}

    The over-representation of NZE lexical items in the Comments of \texttt{r/Philippines} warrants deeper investigation. We inspected the top five NZE dialect features in the Comments of \texttt{r/Philippines}: \textit{mo}, \textit{pa}, \textit{para}, \textit{eh}, and \textit{po}. The over-representation of these features points to the unique challenges posed by borrowings and loanwords in some dialect contexts. Both te reo Māori and Tagalog belong to the Austronesian language family, resulting in shared lexical roots and structural overlap. We compare the token frequencies ($n$) and mean relative frequencies ($\mu_{i}$) of these entries in \texttt{r/newzealand} and \texttt{r/Philippines} in Table \ref{tab:wiktionary_compare}. The unexpectedly high relative frequency of NZE features in the Comments of \texttt{r/Philippines} appears to stem from orthographic homonyms shared between Tagalog particles and te reo Māori loanwords in NZE. The presence of borrowings and loanwords is closely linked to the challenge of multi-word expressions. As presented in Table \ref{tab:te_reo_entries}, a number of multi-word expressions from te reo Māori have been borrowed into NZE. One example, \textit{te reo} (`language'), has a high rate of occurrence in both Submission Posts ($\mu_{i}=11.8$) and Comments ($\mu_{i}=3.8$). Further exploration is required to determine the role of language contact in the development of dialect-responsive resources.

    \begin{table}
        \centering
        \footnotesize
        \begin{tabularx}{\columnwidth}{l *{2}{>{\centering\arraybackslash}X}}
            \toprule
            Entry & Submissions & Comments \\
            \midrule
            \textit{biddi biddi} & - & - \\ 
            \textit{hei tiki}  & - & 42 \\
            \textit{ka pai} & 2 & 619 \\
            \textit{kapa haka} & 8 & 415 \\
            \textit{kia ora} & 121 & 2,645 \\
            \textit{kohanga reo} & - & 90 \\
            \textit{mako shark} & - & 6 \\
            \textit{moho pereru} & - & - \\
            \textit{te ao māori} & 2 & 307 \\
            \textit{te reo} & 155 & 11,421 \\
            \textit{tena koe} & 1 & 58 \\
            \textit{tino rangatiratanga} & 3 & 571 \\
            \textit{whare wānanga} & - & 22 \\
            \bottomrule
        \end{tabularx}
        \caption{\label{tab:te_reo_entries} Total number of multi-word expressions (borrowings from te reo Māori) in \texttt{r/newzealand} from Submission Posts and Comments.}
    \end{table}

\paragraph{Register Effects in Social Media Language} 

    Though not the primary focus of this paper, the divergence in lexical alignment between Submission Posts and Comments underscores the crucial role of register in geo-referenced corpus analysis. As a register, Submission Posts may prompt users to signal regional context through lexis within the constraints of a limited word count. This self-imposed behaviour may have contributed to the close alignment between national dialect features and their corresponding subreddits, particularly for unigrams. Conversely, Comments represent interactive, conversational discourse. In this register, short unigram tokens and cross-linguistic particles (such as the Tagalog homonyms that overlap with te reo Māori borrowings in NZE) dominate frequency counts. This pattern also holds for multi-word expressions, as observed in Tables \ref{tab:mwe_entries} and \ref{tab:te_reo_entries}, which exhibited higher token frequencies in Comments than in Submission Posts.
    
\section{Conclusion}
    This work addresses the challenge of building geographically responsive NLP tools by benchmarking Wiktionary against the OED and evaluating its performance on real-world Reddit data. Overall, our empirical results yield two main insights for computational lexicography and corpus design that allow us to address our original research questions.

    \paragraph{RQ1: How does Wiktionary's regional English coverage compare to traditional lexicons such as the OED?}

    Across the six Inner-Circle varieties, the total sizes of the two lexicons are remarkably comparable: Wiktionary contains 33,043 entries, compared with the OED's 37,679 entries (a 13.1\% difference). The OED is more comprehensive for American and British English; however, Wiktionary has greater coverage of all other varieties of English.

    \paragraph{RQ2: Can Wiktionary-derived features identify regional dialect usage across social media text registers?}

    Wiktionary-derived features can accurately identify regional dialect usage on social media, but only when register context is taken into account (e.g., prioritising titles over comment threads) and when text-preprocessing pipeline rules preserve character-level diacritics and filter out cross-linguistic homonyms.
    
    \paragraph{} Wiktionary serves as a viable, highly responsive complement to curated dictionaries for English varieties. Some of the challenges encountered in evaluating and validating this crowdsourced dialect lexicon illustrate broader issues associated with the use of digital corpora, particularly the influence of language contact in dialects and the effects of register within a corpus. We encourage the use of crowdsourced dialect lexicons to supplement existing corpora, but not to replace them. Future work should evaluate the performance of crowdsourced language tools and resources across different language registers in low-resource contexts. There is also scope to explore how crowdsourced dialect lexicons can be used to evaluate or fine-tune LLMs.

\section*{Limitations}

    We acknowledge that there are differences in the editorial processes between institutional lexicons, like the OED, and the crowdsourcing approach of Wiktionary. Most prominently, crowdsourcing relies on the bottom-up tagging of entries, which differs from top-down lexicographical categorisation. We therefore emphasise that the OED and Wiktionary are not direct like-for-like comparisons, but rather complementary resources. We also recognise that this paper is exploratory and relies heavily on descriptive analysis. As discussed in the conclusion, there is further scope to evaluate crowdsourced lexicons under experimental conditions. Additionally, we recognise that paralinguistic factors have not been fully addressed in the present study, an illustration of which was observed in short unigram tokens for New Zealand English.
    
    Although Wiktionary lists entries such as \textit{mo} (`moustache') and \textit{po} (`a chamber pot') under NZE, these items overlap orthographically with high-frequency Tagalog words. In Tagalog, \textit{mo} serves as a second-person singular possessive pronoun and \textit{po} as a formal honorific. Conversely, in te reo Māori, these terms are written with diacritics as \textit{mō} (a preposition indicating future possession or purpose) and \textit{pō} (`night' or `darkness'). In another example, the unaccented form \textit{pa} functions in Tagalog as an aspectual marker (`still, yet') or an informal noun (`father'), whereas \textit{pā} denotes a `fortified settlement' in te reo Māori (although recorded without the tohutō in Wiktionary). Likewise, \textit{para} designates the `king fern' (Marattia salicina) in te reo Māori but functions in Tagalog as a preposition (`for, in order to'), interjection (`stop'), or verb particle (`seems'). These examples highlight the importance of accounting for cross-linguistic and multilingual contexts when using unstructured natural language data.

    Finally, a key limitation of this study lies in the structural constraints of using country-level subreddits as proxies for geo-referenced language data. Membership in these place-based communities does not guarantee that subscribers physically reside in or are linguistically representative of the target speech communities. However, a similar limitation exists for other social media platforms, such as X \citep{sloan_who_2015}, none of which are fully representative. Online communities present inherent demographic imbalances; for instance, \texttt{r/usa} holds a relatively small membership base (approximately 119,000) relative to the population of the United States, preventing it from serving as a fully representative sample of American English speakers. More broadly, these demographic and sampling biases are not unique to Reddit, but represent an inherent constraint of all web-derived corpora used in dialectology and natural language processing.

\section*{Ethics Statement}

    The rapid rise of proprietary large language models developed by `frontier' AI laboratories has coincided with a notable decline in the digital commons \citep{stalder_decline_2026}. Consequently, the primary purpose of this study is not to advocate for the over-exploitation of Wiktionary as one of the few remaining open, crowdsourced language resources. Nor does this paper suggest that the skilled labour of professional lexicographers can or should be outsourced to volunteer communities on open platforms. Rather, this paper illustrates the evolving role of lexicography in the digital age, demonstrating how crowdsourced lexicons can actively support the development of fair, representative, and equitable language technologies \citep{lew_dictionaries_2024}.

\section*{Acknowledgements}

    This research originated during the author's doctoral studies at Toi Hangarau | Geospatial Research Institute, Te Whare Wānanga o Waitaha | University of Canterbury (UC). The author thanks supervisory committee members Dr Benjamin Adams (UC), Dr Jonathan Dunn (University of Illinois Urbana-Champaign), Prof. Jen Hay (UC), and Dr Kong Meng Liew (UC) for their guidance and feedback. This publication was supported by a Postdoctoral Fellowship from Te Pūnaha Matatini, New Zealand's Centre of Research Excellence for Complex Systems. The author also extends sincere thanks to the programme chairs, organising committee, and four anonymous reviewers for their constructive feedback on earlier versions of this manuscript.


\bibliography{custom}

\appendix

    \begin{figure*}[t]
        \centering
        \includegraphics[width=\linewidth]{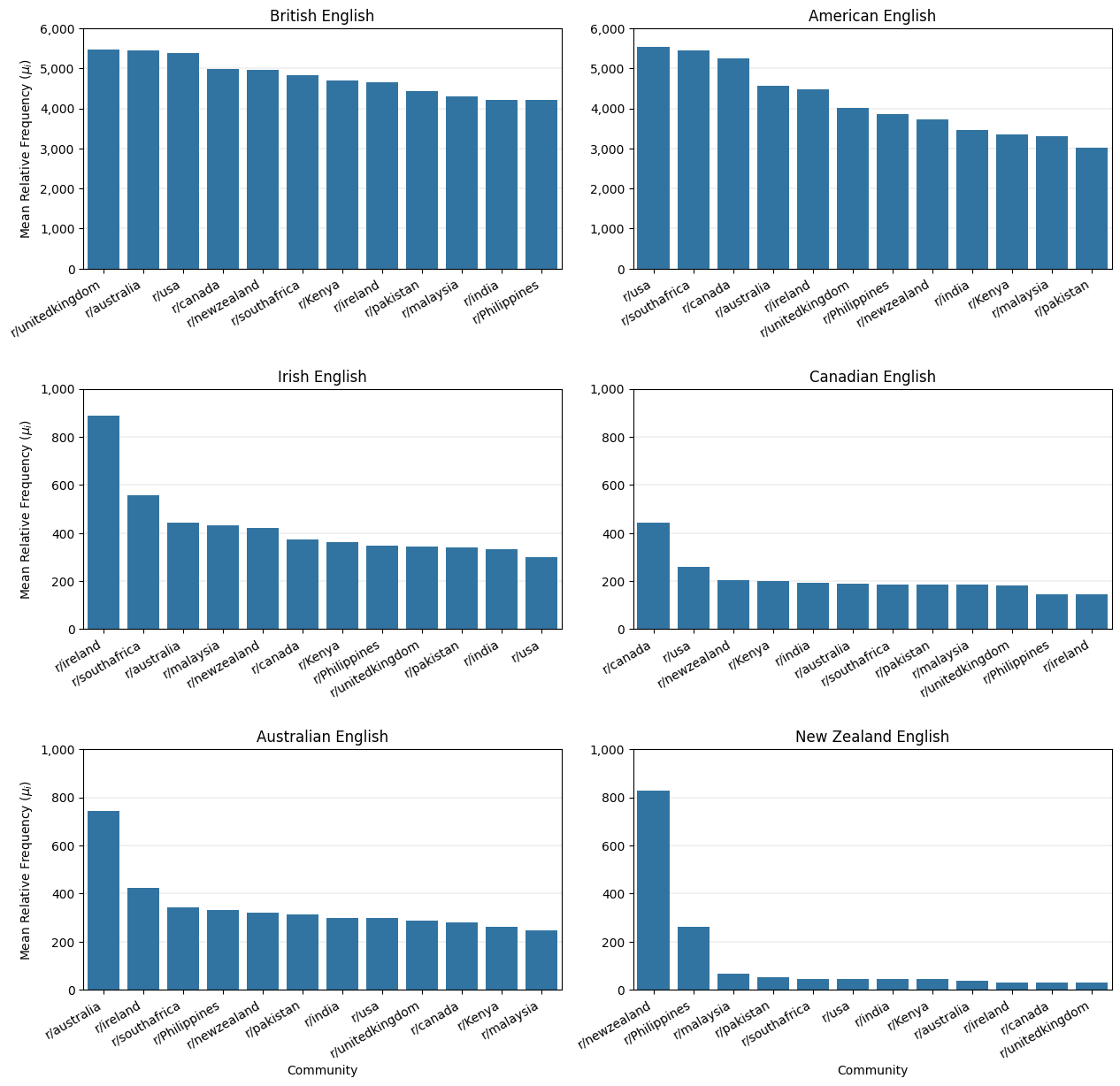}
        \caption{\label{fig:title_lexical} Relative frequencies of lexical dialect features associated with the six Inner-Circle varieties of English across twelve Reddit communities in Submission Posts.}
    \end{figure*}

    \begin{figure*}[t]
        \centering
        \includegraphics[width=\linewidth]{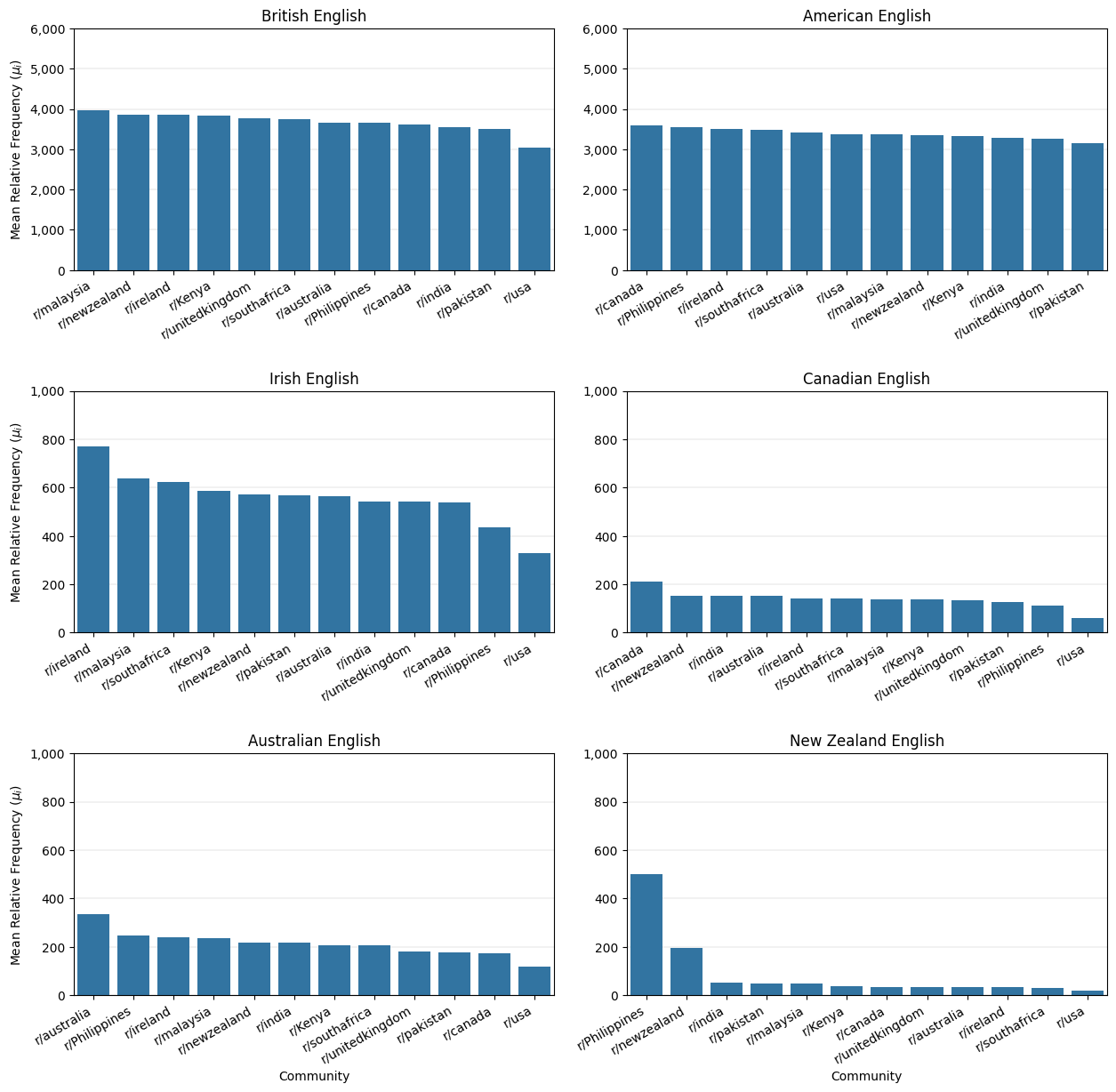}
        \caption{\label{fig:comment_lexical} Relative frequencies of lexical dialect features associated with the six Inner-Circle varieties of English across twelve Reddit communities in Comments.}
    \end{figure*}

\end{document}